\documentclass{article} 
\usepackage{iclr2026_conference,times}
\iclrfinalcopy

\usepackage{amsmath,amsfonts,bm}

\def\eqref#1{equation~\ref{#1}}

\def\1{\bm{1}}

\DeclareMathAlphabet{\mathsfit}{\encodingdefault}{\sfdefault}{m}{sl}
\SetMathAlphabet{\mathsfit}{bold}{\encodingdefault}{\sfdefault}{bx}{n}

\usepackage[utf8]{inputenc} 
\usepackage[T1]{fontenc}    
\usepackage{hyperref}       
\usepackage{url}            
\usepackage{booktabs}       
\usepackage{amsfonts}       
\usepackage{pifont}
\usepackage{nicefrac}       
\usepackage{microtype}      
\usepackage{xcolor}         
\usepackage{multirow}
\usepackage{bbding}
\usepackage{mathtools}
\usepackage{utfsym}
\usepackage{graphicx}
\usepackage{listings}
\usepackage{amssymb}
\usepackage{pifont}
\usepackage{algorithm}
\usepackage{wrapfig}
\usepackage{multicol}
\usepackage{algorithmicx}
\usepackage{algpseudocode}
\usepackage{fancyhdr}
\usepackage{makecell}
\usepackage{pbox}

\title{EMAC+: Embodied Multimodal Agent for Collaborative Planning with VLM+LLM}

\author{%
Shuang Ao\textsuperscript{1}, Flora D. Salim\textsuperscript{1}, Simon Khan\textsuperscript{2}\\
University of New South Wales, Sydney\textsuperscript{1}; \\
Air Force Research Laboratory\textsuperscript{2}\\
\texttt{\{shuang.ao, flora.salim\}@unsw.edu.au}, \texttt{simon.khan@us.af.mil}
}

\begin{document}

\maketitle

\begin{abstract}
Although LLMs demonstrate proficiency in several text-based reasoning and planning tasks, their implementation in robotics control is constrained by significant deficiencies: (1) LLM agents are designed to work mainly with textual inputs rather than visual conditions; (2) Current multimodal agents treat LLMs as static planners, which separates their reasoning from environment dynamics, resulting in actions that do not take domain-specific knowledge into account; and (3) LLMs are not designed to learn from visual interactions, which makes it harder for them to make better policies for specific domains. In this paper, we introduce EMAC+, an Embodied Multimodal Agent that collaboratively integrates LLM and VLM via a bidirectional training paradigm. Unlike existing methods, EMAC+ dynamically refines high-level textual plans generated by an LLM using real-time feedback from a VLM executing low-level visual control tasks. We address critical limitations of previous models by enabling the LLM to internalize visual environment dynamics directly through interactive experience, rather than relying solely on static symbolic mappings. Extensive experimental evaluations on ALFWorld and RT-1 benchmarks demonstrate that EMAC+ achieves superior task performance, robustness against noisy observations, and efficient learning. We also conduct thorough ablation studies and provide detailed analyses of success and failure cases. Code is available at: \href{https://anonymous.4open.science/r/EMAC-Embodied-Multimodal-Agent-for-Collaborative-Planning-with-VLM-LLM-50B2/}{EMAC+ code.}
\end{abstract}

\section{Introduction}
\label{sec:intro}

Recent advancements in Large Language Models (LLMs) have transformed problem-solving in text-centric domains~\citep{Zhu2023MiniGPT4EV, Shinn2023ReflexionLA, Shridhar2020ALFWorldAT, Wang2023DescribeEP}, facilitating complex planning, reasoning, and tool use via human directives. Nonetheless, implementing LLM-based agents for complex tasks~\cite{Jiang2023VIMARM} in real-world scenarios~\cite{Huang2023VoxPoserC3, Mu2023EmbodiedGPTVP, Sumers2023DistillingIV} remains an open problem. Current multimodal agents typically treat LLMs as static planners, using them solely to generate symbolic instructions executed by predefined visual modules~\cite{Driess2023PaLMEAE, Dai2023InstructBLIPTG, Li2023BLIP2BL}. This approach neglects dynamic environmental feedback, hindering adaptability and limiting real-time refinements to action plans. 

Existing methodologies further exacerbate this disparity by regarding LLMs as static, frozen planners, thus separating their textual reasoning from environment dynamics. Models such as SayCan~\cite{Ahn2022DoAI} or Code-as-Policies~\cite{Liang2022CodeAP} exemplify this limitation, utilizing LLMs to generate symbolic action sequences without accounting for significant obstacles like obscured objects, kinematic constraints, or sensor noise. This static nature prevents LLMs from adapting based on interaction histories or updating their understanding of environment-specific constraints like torque limits or object occlusion patterns. Consequently, LLM-based agents frequently propose impractical actions due to a lack of situational awareness and real-time adaptability.

One challenge is that LLM-based agents cannot be directly deployed in a visual world. While LLMs demonstrate impressive abilities in textual interactions, including planning, reflection, and reward shaping, their effectiveness drastically diminishes when directly applied to visual environments. Vision-Language Models (VLMs) have emerged to address this limitation by aligning LLM reasoning with visual inputs. However, existing VLMs predominantly rely on static image-text pair alignments during pretraining, which restricts their ability to understand dynamic visual environments effectively. Even advanced models such as GPT-4V~\cite{2023GPT4VisionSC} struggle in embodied scenarios like ALFWorld, particularly in zero-shot settings~\cite{Shridhar2020ALFWorldAT}. In such cases, these models typically default to leveraging linguistic priors derived from detected objects rather than adapting plans based on task-specific visual dynamics and interactions.

To bridge the gap of deploying LLM in the visual world while maintaining its strengths of both LLM and VLMs, in this paper, we introduce an \textbf{E}mbodied \textbf{M}ulti-modal \textbf{A}gent for \textbf{C}ollaborative planning with LLM \textbf{+} VLM \textbf{(EMAC+)}. EMAC+ allows continuous adaptation of high-level textual action plans based on real-time visual execution feedback. Unlike existing methods, this collaboration empowers the LLM to internalize environment-specific dynamics and improve its domain-specific comprehension, thus generating more accurate and feasible plans for complex robotic tasks. Our approach addresses critical limitations in previous static multimodal methods by providing dynamic, interactive updates to action planning through direct visual feedback. Moreover, with its understanding of physical dynamics, EMAC+ can also generate precise control actions in robotics tasks. While recent VLA models like PaLM-E~\cite{Driess2023PaLMEAE} also output low-level actions, our framework differs in two key ways: \textbf{(1) A collaborative scheme for LLM + VLM}: EMAC+ first uses the prior knowledge from an LLM expert to predict the action sequences. The VLM agent will execute the planned actions to interact with the visual environment, and provide feedback for the LLM expert, including the knowledge of the history trajectories and the environment dynamics.  \textbf{(2) Domain-aware planning}: Due to VLM feedback, our LLM expert will possess both contextual knowledge and practical experience of the agent's execution. In contrast to PaLM-E’s open-loop execution, our LLM systematically refines its plans based on feedback from the VLM. This closed-loop synergy ensures that both models change simultaneously, with the LLM changing to suggest actions that work with the VLM's control abilities.

\begin{figure}[tbh]
    \centering
    \vspace{-1em}
    \includegraphics[width=.9\textwidth]{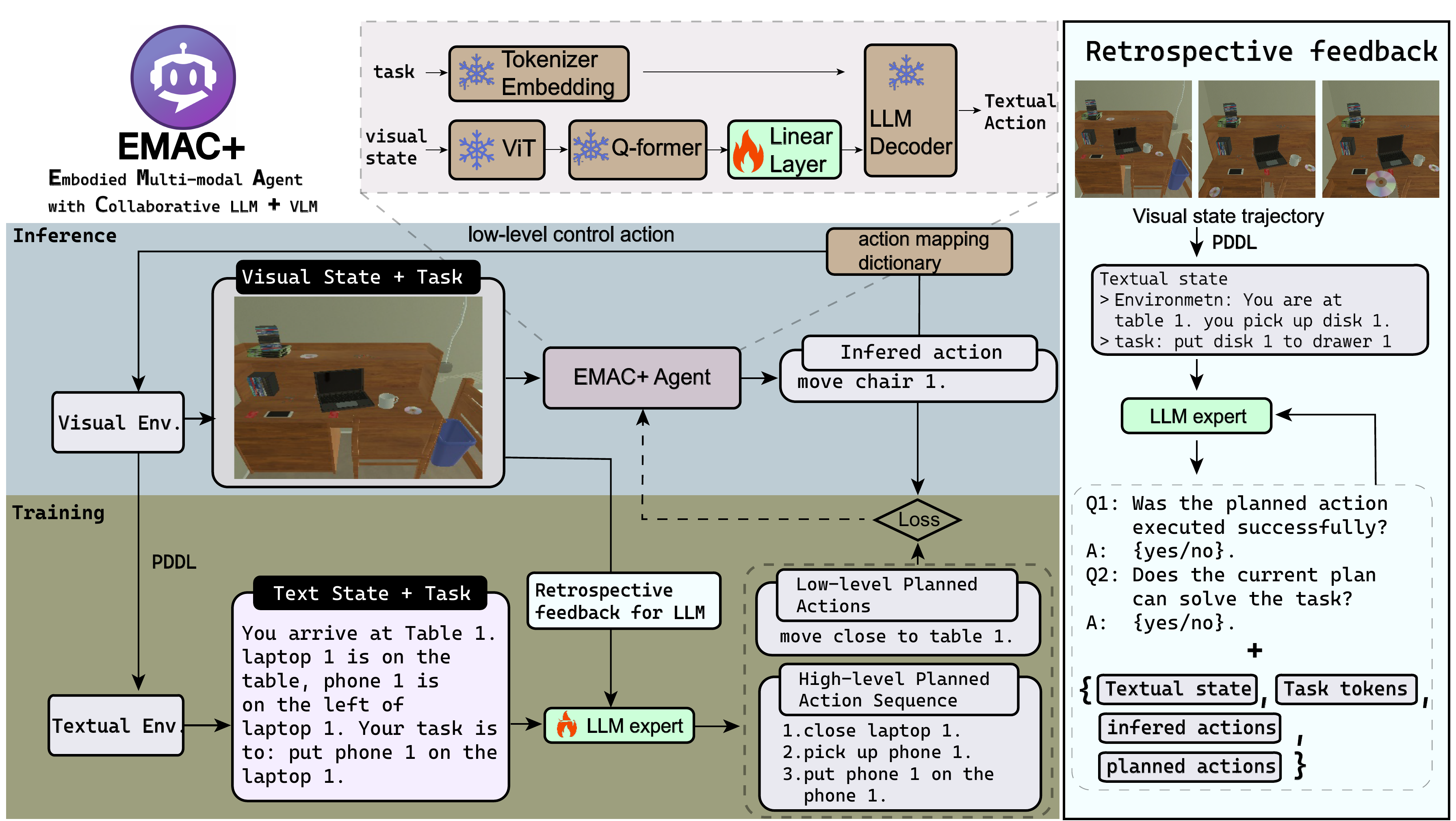}
    \vspace{-.5em}
    \caption{\footnotesize\textbf{EMAC+: Embodied Multi-modal Agent for Collaborative Planning with LLM + VLM.} EMAC+ takes a task instruction and pixel observations as input to plan the sequence of actions to complete tasks. Moreover, EMAC+ can also be directly deployed into the task/environment, which requires low-level control actions for the interaction. EMAC+ utilizes both the prior knowledge from LLM expert and the domain-specific knowledge from the environment dynamics.}
    \vspace{-1em}
    \label{fig:structure}
\end{figure}

We validate EMAC+ in two benchmarks: ALFWorld~\cite{Shridhar2020ALFWorldAT}, which includes numerous tasks in both visual and textual environments, and RT-1~\cite{Brohan2022RT1RT}, a real-world robotic platform for manipulation. Experiments demonstrate that EMAC+ achieves better performance in both benchmarks with a higher learning efficiency. Meanwhile, EMAC+ demonstrates enhanced generalization capability for out-of-distribution tasks, and exhibits more consistent performance amid environment disturbances. These results highlight three key contributions: (1) a bidirectional training paradigm that harmonizes symbolic reasoning with vision-driven control, enabling the LLM to learn environment dynamics through the VLM’s feedback; (2) an embodied agent that directly generates low-level actions from pixels while retaining the interpretability of language-guided planning; and (3) validation of a co-adaptive LLM-VLM framework in both simulated and real physical environments. Our research propels the boundaries of LLMs as adaptive controllers, facilitating the development of resilient, generalizable agents that acquire knowledge through visual interactions while preserving the natural language-driven reasoning. 

\section{Related Works}
\label{sec:relatedworks}

A key hurdle for LLM/VLM agents is their inability to generalize beyond training distributions. For example, CLIPort~\cite{Shridhar2021CLIPortWA} demonstrates strong performance on tabletop tasks but fails with novel object configurations or lighting conditions. Similarly, VIMA~\cite{Jiang2023VIMARM} highlights that VLMs trained on static image-text pairs struggle to adapt to dynamic, partially observable RL environments. Recent work by ~\cite{lin2025sim} identifies a ``sim-to-real perception gap'' where agents overfit to simulated visuals and fail under real-world sensory noise. 
LLM-generated plans (e.g., ``pick up the tool'') often lack precision for low-level control, leading to execution failures. Efforts like Distilled Behavior Cloning~\cite{Liang2024MonitoringAC} attempt to mitigate these issues by distilling LLM knowledge into RL policies, but they sacrifice the flexibility of direct language grounding. Due to the nature of language, the textual descriptions of pixel observations are prone to losing important details for robotic control. Unlike these methods, EMAC+ integrates real-time visual feedback into a collaborative loop between LLM and VLM, allowing continuous refinement of plans based on dynamic observations. This interactive process enables EMAC+ to adapt flexibly to novel scenarios and environmental perturbations, thereby significantly mitigating the generalization issues traditional static-alignment models face. Consequently, EMAC+ can effectively translate high-level textual instructions into precise, executable actions suitable for robust robotic control tasks.

Aligning LLMs and VLMs requires bridging semantic and embodiment gaps. Semantically, LLMs reason in abstract language, while VLMs map pixels to text, creating a mismatch in representational granularity. For instance, LLaVA~\cite{Liu2023VisualIT} aligns LLMs with VLMs for visual QA but lacks mechanisms to translate answers into actions. Architectures like PaLM-E~\cite{Driess2023PaLMEAE} embed both modalities into a shared latent space but struggle with real-time policy updates during RL training. More specifically, the reward design in alignment is challenging, i.e., LLM-generated rewards (e.g., the correctness of step-by-step plans) often misalign with environment-specific success signals~\cite{Wang2023DescribeEP}. Then, VLMs process static frames, while LLMs generate sequence-based plans, leading to incoherent action sequences in long-horizon tasks. On the other hand, jointly fine-tuning LLMs and VLMs risks catastrophic forgetting of pre-trained knowledge. Moreover, VLMs~\cite{Li2023MultimodalFM} enable plans to be conditioned on visual inputs that are converted into linguistic descriptions or token embeddings aligned with LLMs~\cite{Ahn2022DoAI,Gao2023PhysicallyGV,Huang2023VoxPoserC3, Helmert2006TheFD}. Nevertheless, current foundation models are typically pretrained on static text or text-image datasets~\cite{Driess2023PaLMEAE, Mu2023EmbodiedGPTVP, Yang2023OctopusEV, Zhong2024PolicyIU}, which may hinder their ability to adapt to the world's changes. Due to the nature of language, the textual descriptions of pixel observations are prone to losing important details for robotic control. EMAC+ mitigates the issue of imprudent planning activities resulting from absent critical knowledge by analyzing both short-term and long-term historical data, encompassing the historical trajectories of the interaction between the VLM and its environment. Simultaneously, by analyzing failure paths, the LLM agent will avoid the forgetting problems and actively comprehend the environmental physical knowledge, enhancing the planning quality.

\vspace{-.5em}
\section{Methodology}
\vspace{-.5em}
Fig.~\ref{fig:structure} illustrates the main idea of our multi-modal agent, who trained by the interactive knowledge between LLM and VLM. Our agent is built on a modularized VLM, which can take pixel observations as input and generate low-level control actions. We follow ~\cite{Yang2023EmbodiedMA} to construct a parallel text world for providing EMAC+ with step guidance to achieve this and overcome the associated challenges. Moreover, we design an interactive training scheme for the LLM expert to allow it to have a clear understanding of the gap between the text world and the real visual environment.

\vspace{-.5em}
\subsection{Multi-Modal Agent}
\vspace{-.5em}
In visual environments, the multi-modal agent $\pi_\theta$ is designed to process a task instruction $x_{task}$, and the pixel observation $s_v^t$ at each interaction step $t$. $\pi_\theta$ is trained to generate a sequence of \textbf{real control actions $\left\{x_a^t\sim\pi_\theta(\cdot|x_{task},s_v^t)\right\}$} to control the agent/robot to complete the task efficiently. To achieve this, inspired by recent works~\citep{Dai2023InstructBLIPTG, Li2023MultimodalFM, Li2023BLIP2BL, Liu2023VisualIT, Yang2023EmbodiedMA, Zhu2023MiniGPT4EV}, we design the VLM-based agent of five components: (1) a ViT to encode $s_v$ into visual embeddings, (2) a querying transformer (Q-Former) tailored to extract the most relevant visual features via the cross-attention between the visual embeddings and query tokens, (3) a linear projection layer to align visual features to text embeddings, (4) an LLM decoder taking the concatenation of the instruction tokens and the output of the linear projection layer to autoregressively generate the action $x_a$, (5) a pre-trained action dictionary mapping the textual actions to control actions. In our model, we adopt the pre-trained ViT, Q-former, and LLM from InstructBLIP~\citep {Dai2023InstructBLIPTG} and keep them frozen at the finetuning stage. Note that we only update the linear projection layer as shown in Fig.~\ref{fig:structure}. To enable our multi-modal agent to work in different tasks, e.g., manipulation tasks, which require the model to generate control actions. Given the previous success of PaLM-E~\citep{Driess2023PaLMEAE}, it is reasonable to assume that we have access to policies that can perform low-level skills from some (small) vocabulary as long as the multi-modal agent consists of a sequence of such skills. Thus, we adopt an action mapping dictionary that contains low-level skills (e.g., ``rotate the arm with $90^{\circ}$''), which maps such skills to control actions. \looseness-1

Nonetheless, there are persistent issues that require our attention, as outlined in Sec.~\ref{sec:intro}. While the pre-trained VLM~\citep{Dai2023InstructBLIPTG, Li2023MultimodalFM, Li2023BLIP2BL} is functional, it is suboptimal, as the pre-training process restricts the model to focus solely on the static alignment of image-text pairs, and the trained agent cannot reason about the dynamics of the environment. 
For example, even GPT-4V, a SOTA VLM, fails to accomplish tasks in an embodied visual environment that contains complex dynamics and object relationships (e.g., ALFWorld~\citep{Li2023MultimodalFM} in our experiments). More specifically, the primary reason for the failure of VLM agents is their reliance on the logical relationships of currently observed objects to create action instructions while neglecting the task instruction and the underlying dynamics of the environment. The strength of LLM lies in its ability to identify similarities and commonalities despite variations in details across multiple environments. Furthermore, finetuning a pre-trained VLM on a pre-collected dataset is still suboptimal due to the variability of environments and tasks~\citep{Brohan2023RT2VM}, the absence of large-scale expert annotations~\citep{Padalkar2023OpenXR}, and the challenges arising from distribution shift issues~\citep{Levine2020OfflineRL}. \looseness-1

A natural solution to the above challenges is using reinforcement learning environment feedback (RLEF), where the reward indicates the completion of sub-goals. Therefore, it is straightforward to consider a collaborative scheme for the LLM expert and VLM agent, where LLM plays the role of high-level planner and VLM as an executor. Meanwhile, LLM can decide whether to re-plan based on the environment feedback. We design a collaborative learning process to make EMAC+ align with the environment dynamics and the expert knowledge from LLM. This brings up two problems: (1) How to obtain an effective expert that EMAC+ can query? (2) How can the LLM expert understand the environment dynamics and learning process?

\begin{wrapfigure}[22]{r}{0.65\textwidth}
\vspace{-2.2em}
\begin{minipage}{0.65\textwidth}
\begin{algorithm}[H]
\footnotesize
\caption{EMAC+: Collaborative VLM+LLM}
\label{alg:design}
\begin{algorithmic}[1]
\State {\bf Initialize:} $i=0$, $\mathcal D \leftarrow \emptyset$, $\mathcal D_{ref} \leftarrow \emptyset$, LLM agent $M_l$;
\State {\bf Input:} max trials $I$, training epochs$I_e$, visual and textual environment $E_v, E_l$, task instruction $x_{task}$, expert agent $\pi_{ref}$, EMAC+ policy $\pi_\theta$, LLM agent memory pool $\mathcal P \leftarrow \emptyset$;
\State Initialize $\pi_\theta$ to $\pi_{ref}$;
\While{$i<I$ or task not completed}
\State Get $\tau_v^i=[x_{task}, s_v^0, x_a^0, \cdots, s_v^T, x_a^T,]$ from $E_v$ with $\pi_\theta$;
\State Get $\tau_l^i=[x_{task}, s_l^0, x_a^0, \cdots, s_v^T, x_a^T,]$ from $E_l$ with $\tau_v^i$;
\State Generate LLM feedback $\mathcal P_i = M_l(\tau_l^i, \tau_v^i)$;
\For{$t=0$ to $T$} \Comment{Dataset Aggregation}
\State Planned action $x^*_a = M_l (\mathcal P, x_{task}, \cdots, x_a^{t-1}, s_l^t)$;
\State $\mathcal D \leftarrow \mathcal D \cup \left\{x_{task}, s_v^t, x_a^t, x_a^*\right\}$;
\State $\mathcal P \leftarrow \mathcal P \cup \left\{x_{task}, s_v^T, x_a^T, x_a^*\right\}$;
\EndFor
\For{$j=0$ to $I_e -1 $} \Comment{Training and Finetuning}
\State Sample a minibatch from $\mathcal D$;
\State Update $\theta$ by Eq.~\ref{equ:loss} with $\pi_{ref}$ and $\tau$;
\EndFor
\State Finetune $\pi_{ref}$ by Eq.~\ref{equ:finetune_loss} with $\mathcal P$;
\State $i\leftarrow i+1;$
\EndWhile
\end{algorithmic}
\end{algorithm}
\end{minipage}
\end{wrapfigure}

\subsection{Text descriptions on visual observations}
A core challenge in aligning vision-language models (VLMs) with large language models (LLMs) for manipulation tasks is bridging the gap between pixel-level observations and the symbolic reasoning required by LLMs. A common way to overcome this gap is to utilize PDDL (Planning Domain Definition Language) to convert raw visual observations into structured textual states~\cite{Yang2023EmbodiedMA, Shridhar2020ALFWorldAT}. This enables the LLM to interpret the environment symbolically while retaining actionable semantics for downstream planning. The translation module integrates key attributes, such as \texttt{Observed Objects, observed relations, inventory, and locations}. Nevertheless, the text derived from pixel observation can solely furnish the LLM with abstract and ambiguous descriptions, merely encapsulating the fundamental information in the environment but failing to convey the physical specifics of the particular environment, task, and comprehensive action descriptions. While textual descriptions suffice for certain jobs (such as alfworld~\cite{Shridhar2020ALFWorldAT} and epic-kitchen~\cite{Damen2018epickitchen}), they are inadequate for robotic control tasks that necessitate greater precision. Therefore, in EMAC+, we set up a retrospective feedback for the LLM agent to build a closed-loop interaction between the LLM agent and the VLM agent. Within this loop, the LLM agent not only has the short-term memory of planned actions, but it will also have the long-term memory of the historical trajectories of the agent in the environment for re-planning, which allows the LLM agent to understand the dynamics of the visual world and change the planned action sequences accordingly.


\vspace{-.5em}
\subsection{Build Collaborative LLM and VLM}
\vspace{-.5em}
Given the parallel text world translated from the visual world, we aim to train a VLM agent $\pi_\theta$ to complete tasks in the visual world. To align the prior knowledge provided by the LLM expert with the VLM, inspired by ~\cite{Yang2023EmbodiedMA}, we firstly formulate it as an imitation learning, and the objective can be written as:
~\begin{equation}\label{equ:imit_loss}
    \theta ^* = \arg\min_{\theta \in \Theta} \mathbb{R}_{\pi_\theta} \left[\mathcal L_{imit} (\pi_\theta(x_a|s_v), x_a^*\right],
\end{equation}
in which $\mathcal L$ is dependent on the specific tasks. In EMAC+, we use DPO~\citep{Rafailov2023DirectPO} loss as its proven superior performance in the cross-entropy on aligning models with expert preferences within the discrete language space. Thus, we rewrite Eq.~\ref{equ:imit_loss} as:

\begin{align}\label{equ:loss}
& \theta^*=\underset{\theta \in \Theta}{\arg \min }-\mathbb{E}_{\pi_\theta}\left[\mathcal{L}_{imit}(\pi_\theta, \pi_{ref}, s_v, x_a, x_a^*)\right], \\
\nonumber& \mathcal{L}_{imit}(\cdot) \triangleq \log \sigma\left(\beta \log \frac{\pi_\theta(x_a^*|s_v)}{\pi_{ref}((x_a^*|s_v)}-\beta \log \frac{\pi_\theta(x_a|s_v)}{\pi_{ref }\left(x_a|s_v\right)}\right),
\end{align}

where $x_a^*$ is the planned action generated by an expert while $x_a$ is generated by $\pi_\theta$, $\sigma$ is the logistic function and $\beta$ is a hyperparameter controlling the deviation from $\pi_{ref}$. We utlize this regularization to prevent the agent from deviating too far from the distribution on which the expert is accurate.

Given a pre-trained VLM agent learned from an expert, we now build a closed-loop collaboration between VLM and LLM. Given a task instruction $x_{task}$ and the textual observation $s_l$ translated by PDDL, the LLM agent first generates a high-level textual action sequence $x_{a,1:N}$ that is expected to complete $x_{task}$ step-by-step. LLM also infers low-level execution $b$ (one step action) given $=x_{a,1:N}$ and $s_l$. VLM agent executes the control actions to interact with the environment based on $b$. After the execution, the environment returns a new pixel observation $s_v$. LLM agent will decide whether it needs to modify $x_{a,1:N}$ to $x^*_{a,1:N}$ based on $s_v$ and the corresponding text state $s_l$. Thus, we collect the trajectories $\mathcal D = \left\{(s_t, g, x_{a,1:N}, x^*_{a,1:N}\right\}$. We fine-tune the LLM with LoRA~\citep{Hu2021LoRALA} for parameter-efficient updates, retaining the LLM’s general knowledge while adapting to environment dynamics as:
\begin{equation} \label{equ:finetune_loss}
    \mathcal L = - \sum_{i=1}^N \log P(x_{a,i}^*|s_t,g,x_{a,<i}^*).
\end{equation}

Thus, we formulate a collaborative framework for EMAC+ (refer to Algorithm~\ref {alg:design}), which involves an LLM as an expert that provides action planning, and a VLM agent executes the low-level actions. Meanwhile, LLM utilizes the environment's feedback to modify the planning and adapt it to the agent's learning process.

\textbf{EMAC+ in control loop} EMAC+ can not only produce textual actions based on the multi-modal sentences as input, it also can be used to generate low-level commands to solve an embodied planning or control task. More specifically, we assume access to the control policy capable of executing low-level skills from a pre-trained (limited) action vocabulary, and an effective plan from EMAC+ should comprise a succession of these abilities. Thus, EMAC+ needs to independently ascertain the available skills based on the environment feedback and the prompt, without any additional mechanisms to restrict or filter its output. Note that the control policies are insufficient for solving long-horizon tasks or understanding complex instructions.
EMAC+ is included in a control loop, where its projected decisions are executed through low-level policies by a robot, generating new observations that allow EMAC+ to replan. In this perspective, EMAC+ can be considered a high-level policy that structures and controls low-level policies.

\vspace{-.5em}
\section{Experiments}
\vspace{-.5em}
\subsection{Experiments Setup}
\vspace{-.5em}
\paragraph{Environments.} We base our experiments on two embodied environments, ALFWorld~\citep{Shridhar2020ALFWorldAT} and RT-1~\citep{Brohan2022RT1RT}. The former environment provides a cross-modality simulation platform that encompasses a wide range of embodied household tasks. RT-1 has three robot environments, including a Task and Motion Planning (TAMP) domain where a robot has to manipulate (grasp and stack) objects, a tabletop pushing
environment, and a mobile manipulation domain. We provide more explanations and details for each environment as follows: \looseness-1

\textbf{ALFWorld.} For each task in ALFWorld, the simulator integrates a visual environment, rendered by the AI2THor~\citep{Kolve2017AI2THORAI}, with a corresponding
textual environment. The textual environment employs PDDL~\citep{Helmert2006TheFD} to translate
each pixel observation from the simulator into an equivalent
text-based observation, and then construct an interactive
world using the TextWorld engine~\citep{Ct2018TextWorldAL}. The
tasks within the benchmark are categorized into six types: ``Pick
\& Place'', ``Clean \& Place'', ``Heat \& Place'', ``Cool \& Place'', ``Look in Light, and Pick Two Objects \& Place''. Each task requires an agent to execute a series of text-based actions, such as ``go to safe 1'', ``open safe 1'', or ``heat egg 1 with microwave 1'', following a predefined instruction. These actions involve navigating and interacting with the environment. A human expert might need more than 30 steps to solve a job in this benchmark that involves interacting with more than 10 objects. This makes it hard for an agent to plan ahead, comply with instructions, and use common sense knowledge. To make the comparison fair, we use the same setting as earlier work and use 134 out-of-distribution (OOD) jobs to test all of the baselines.

\textbf{RT-1.} RT-1 has three domains. We train EMAC+ on expert data from each domain of RT-1. In many cases, this is a sparse amount of data per task. The TAMP problems require a lot of combinatorial analysis of possible plans, and many decision sequences can't be made. EMAC+ must formulate strategies comprising several stages, characterized by intricate decision limits. The multi-object tabletop pushing environment is derived from the publicly accessible Language-Table dataset~\citep{Lynch2022InteractiveLT} and presents challenges due to the presence of several objects, a high cardinality of language, and intricate pushing dynamics. In both the TAMP and Language-Table environments, EMAC+ must analyze the positions of the objects. Merely identifying the things on the table and their approximate relationships is inadequate; a more nuanced understanding of the scene's geometry is essential for task resolution.

\textbf{Baselines.} To verify the effectiveness of the collaborative learning scheme, we compare EMAC+ with several baselines and SOTA agents. We categorize the baselines with three categories: vision models, language models (LLM agents), and vision-language models (VLM agents). For vision models, we pick ResNet-18~\cite{He2015DeepRL} and MCNN-FPN~\cite{He2017MaskR}, which utilize pretrained encoders to extract salient features from each pixel observation. Then, behavior cloning is used to train a Multi-Layer Perceptron (MLP) policy on a demonstration dataset that was already made using the extracted features as background. In contrast to vision models that operate inside a visual environment, language models do analogous tasks in a parallel, text-based framework. Qwen~\citep{Bai2023QwenVLAF} achieves competitive performance on diverse benchmarks and illustrates its strong potential on general tasks. BUTLER~\citep{Shridhar2020ALFWorldAT} utilizes a transformer sequence-to-sequence model augmented with a pointer softmax method~\citep{Glehre2016PointingTU}. This design compiles prior observations as input to produce text-based actions incrementally, one token at a time. GPT-BUTLER~\citep{Micheli2021LanguageMA}, a variation of the GPT-2 model~\citep{Radford2019LanguageMA}, is initially pretrained on a static demonstration dataset and subsequently finetuned with data gathered online. ReAct~\citep{Yao2022ReActSR} employs an innovative method by leveraging LLMs to produce reasoning traces and task-specific actions concurrently. This strategy assists the agent in formulating, monitoring, and revising its action plans interactively. Reflexion~\citep{Shinn2023ReflexionLA} similarly utilizes a large language model, concentrating on the analysis of environment feedback. It retains this reflective text in an episodic memory buffer, augmenting the agent's capacity to refine actions in future trials. DEPS~\citep{Wang2023DescribeEP} rectifies inaccuracies in prior LLM-generated actions by incorporating descriptions of the action execution process and offering self-explanations for the feedback. Furthermore, beyond the single-agent paradigm, WALL-E~\citep{Zhou2024WALLEWA} has SOTA performance on ALFWorld across various LLM-based agents, and AutoGen~\citep{Wu2023AutoGenEN} demonstrates the capability to do a wide array of tasks through the collaboration of numerous LLM agents. We examine a variety of vision-language models, including MiniGPT-4~\citep{Zhu2023MiniGPT4EV}, BLIP-2~\citep{Li2023BLIP2BL}, LLaMA-Adaptor~\citep{Gao2023LLaMAAdapterVP}, and InstructBLIP~\citep{Dai2023InstructBLIPTG}, as agents for engaging with the visual environment. In contrast to pure vision or language models, VLMs are engineered to interpret and synthesize both visual and verbal information, providing a more comprehensive understanding of the environment. We finetune these agents on a pre-collected demonstration dataset to align them with the unique needs of the ALFWorld benchmark. In RT-1, We pick PaLM-E~\citep{Driess2023PaLMEAE}, SayCan~\citep{Ahn2022DoAI}, and PaLI~\citep{Chen2022PaLIAJ} as baselines, since the RT-1 environment requires the agent to generate low-level control actions rather than textual actions. \looseness-1

\vspace{-.5em}
\subsection{ALFWorld Environment}
\vspace{-.5em}
The architectural design of EMAC+ is shown in Fig.~\ref{fig:structure}. The core of the VLM part in EMAC+ employs a Query Transformer (Q-former) to process the visual data. a. This Q-Former extracts features using a frozen ViT encoder. The output of this Q-Former comprises 32 visual tokens, which undergo a linear projection layer before reaching an LLM decoder. In our experiments, we choose LLaMa3 as the LLM expert due to its outstanding performance in reasoning and planning tasks~\citep{Shinn2023ReflexionLA, Wang2023DescribeEP, Wu2023AutoGenEN, Yao2022ReActSR}.  \looseness -1

\begin{table}[tbh]
    \centering
    \footnotesize
    \vspace{-.5em}
    \resizebox{.95\textwidth}{31mm}{
    \begin{tabular}{lccccccccc}
    \toprule
    \hline
    \multirow{2}{5em}{\textbf{Agent}} &\multirow{2}{4em}{\textbf{Visual Env.}}&\multirow{2}{4em}{\textbf{Textual Env.}}& \multicolumn{7}{c}{\textbf{Success Rate}} \\
    \cline{4-10} && & Avg. & Pick & Clean & Heat & Cool & Look & Pick2 \\
    \midrule
    ResNet- 18* & \checkmark & \ding{55}
    & 0.06 & - & - & - & - & - & - \\
        MCNN-FPN* & \checkmark & \ding{55}
        & 0.05 & - & - & - & - & - & - \\
        \midrule
        \textit{LLMs Agent}\\
        BUTLER$*$ & \ding{55} &\checkmark & 0.26 (-) & 0.31 (-)  & 0.41 (-) & 0.60 (-) & 0.27 (-) & 0.12 (-) & 0.29 (-) \\
        GPT-BUTLER & \ding{55} &\checkmark & 0.69(18.8) &0.62(18.4) &0.81(18.4)& 0.85(12.7) &0.78(15.6) &0.50(24.4) & 0.47(26.6) \\
        ReAct &  \ding{55} &\checkmark & 0.54(20.6) & 0.71 (18.1) & 0.65 (18.8) & 0.62(18.2) & 0.44 (23.2) & 0.28 (23.7) & 0.35 (25.5) \\
        Reflexion& \ding{55} &\checkmark & $\mathbf{ 0.91}$ (18.7) & $\mathbf{0 . 9 6}(17.4)$ & $\mathbf{1 . 0 0}(17.0)$ & 0.81 (19.4) & 0.83 (21.6) & 0.94 (16.9) & $\mathbf{0 . 8 8}(21.6)$ \\
        DEPS$*$ & \ding{55} &\checkmark & 0.76(-) & 0.93(-)& 0.50(-)& 0.80(-) &1.00(-) &1.00(-) &0.00(-) \\
        AutoGen$*$ & \ding{55} &\checkmark &  0.77(-) & 0.92(-) & 0.74(-) & 0.78(-) & 0.86(-) & 0.83(-) & 0.41(-) \\
        AdaPlanner
        & \ding{55} &\checkmark & 0.91(-) & 1.00(-) & 1.00(-) & 0.89(-) & 1.00(-) & 0.97(-) & 0.47(-)\\
        Qwen2-Chat & \ding{55} &\checkmark  & 0.67(-) & 0.62 (-) & 0.75 (-) & 0.68(-) & 0.61(-) & 0.78 (-) & 0.52 (-)\\
        WALL-E & \ding{55} &\checkmark & 0.95(-) & 1.00(-) & 0.97(-) & 1.00(-) & 0.86(-) & 0.83(-) & 0.95(-)\\
        \midrule
        \textit{VLMs Agent}\\
        MiniGPT-4 & \checkmark & \ding{55}
        & 0.16(26.9) & 0.04 (29.0) & 0.00 (30.0) & 0.19 (26.3) & 0.17 (26.7) & 0.67 (17.7) & 0.06(28.9)\\
        BLIP-2 & \checkmark & \ding{55}
        & 0.04 (29.5) & 0.00 (30.0) & 0.06(29.3) & 0.04(29.9) & 0.11 (28.2) & 0.06(29.2) & 0.00 (30.0) \\
        LLaMA-Adapter & \checkmark & \ding{55}
        & 0.13 (27.5) & 0.17 (26.4) & 0.10 (28.6) & 0.27 (24.2) & 0.22 (26.7) & 0.00 (30.0) & 0.00 (30.0)\\
        InstructBLIP& \checkmark & \ding{55}
        & 0.22 (26.2) & 0.50 (21.5) & 0.26(25.0) & 0.23(27.2) & 0.06(28.9) & 0.17 (26.8) & 0.00 (30.0)\\
        EMMA$*$ & \checkmark & \ding{55}  & 0.82 (19.5) & 0.71 (19.3) & \textbf{0.94} (17.5) & 0.85(19.6) & 0.83(19.9) & 0.88 (19.6) & 0.67 (22.4)\\
        Qwen2.5 & \checkmark & \ding{55}  & 0.72 (18.4) & 0.61 (16.3) & 0.85 (18.5) & 0.73(19.7) & 0.75(20.2) & 0.80 (21.2) & 0.55 (26.4)\\
        \textbf{EMAC+} (Ours) & \checkmark & \ding{55} & 0.88 (17.5)& \textbf{0.79} (17.3) & 0.93 (15.5)& \textbf{0.90} (18.8) & \textbf{0.90} (18.3) & \textbf{0.88} (16.6) & \textbf{0.74} (21.0)\\
        \midrule
        \multirow{2}{8em}{\textbf{Human  Performance*}}& \multirow{2}{.5em}{\checkmark} & \multirow{2}{.5em}{\ding{55}} &\multirow{2}{1.7em}{\textbf{0.91}}& \multirow{2}{.5em}{-} & \multirow{2}{.5em}{-} &  \multirow{2}{.5em}{-} &  \multirow{2}{.5em}{-} &  \multirow{2}{.5em}{-} &  \multirow{2}{.5em}{-} \\
        \\
        \hline
        \bottomrule
    \end{tabular}}
    \vspace{-.5em}
    \caption{\footnotesize\textbf{Comparison with the SOTA agents in ALFWorld.} $*$ - reported in the previous work. The highest scores for each task in the same type of environment are highlighted in bold. \checkmark and \ding{55} denote the corresponding environment used/not used to deploy the agent. The average interaction steps are given in parentheses.}
    \label{tab:alfworld}
    \vspace{-.5em}
\end{table}

We find the EMAC+ outperforms all baselines on the ALFWorld benchmark with a lower training cost. We compare EMAC+ with 13 other representative agents on both textual and visual environments and report the results in Table.~\ref{tab:alfworld}. We evaluate two main metrics: the success rate, which is the percentage of successful trials, and the average number of interaction steps needed to finish the task. A lower number of steps means the task was completed more quickly. EMAC+ has exceptional performance across both criteria, surpassing all VLM agents in visual contexts. Additionally, EMAC's performance is significantly better than that of VM agents, highlighting the importance of having prior information built in for VLMs.
Interestingly, EMAC+'s performance is analogous to that of LLM agents utilizing precise semantic representations of visual observations. We attribute this to the training strategy, which involves imitating from an LLM expert agent instead of learning from scratch in a purely visual setting. Consequently, EMAC+ stands out as a VLM agent that substantially surpasses SOTA LLM agents, such as AutoGen~\citep{Wu2023AutoGenEN} and DEPS~\citep{Wang2023DescribeEP}.

We found that \textbf{EMAC+ is more robust to noisy observations than LLM-only agents} (Figure.~\ref{fig:enter-label}, Left). Although LLM agents demonstrate an elevated success rate with fewer interaction steps in textual settings, as shown in Table.~\ref{tab:alfworld}, we assume that this enhanced performance mostly depends on their accurate semantic abstraction of the environment. Nonetheless, this abstraction may prove impractical in real-world applications. To validate this assumption, we established a more practical environment in which observations are intentionally distorted at a designated noise rate. In the next step, we test how well EMAC+ and Reflexion, a cutting-edge LLM agent, can handle these noisy observations. We crop, enlarge, and substitute the original observation with a random segment of the visual input to produce noisy observations. Similarly, we replace random tokens in the textual observations with arbitrary alternatives. With increasing interference, EMAC+ demonstrates significantly greater stability compared to the state-of-the-art LLM agent. The encoder in VLM effectively filters noise, but the textual noise (replacement tokens) is directly handled by LLM. Conversely, it is seen that EMAC+ exhibits greater resistance to interference than EMMA, a distinction ascribed to the capability of our LLM expert to dynamically modify the planned actions based on the environment feedback and the current agent state.

\vspace{-.5em}
\subsection{RT-1 Environment}
\vspace{-.5em}
For RT-1 robot tasks, we run our experiments in a simulator on the RT-1 codebase~\citep{Brohan2022RT1RT}. We report the results on motion planning tasks, table pushing tasks and manipulation tasks in Figure.~\ref{tab:RT1-TAMP} - Figure.~\ref{tab:RT1-manipulation}. 

\textbf{RT-1 TAMP environment.} Table.~\ref{tab:RT1-TAMP} reports the success rates and VQA performance in the TAMP environment. Note that the LLM in baseline methods is frozen (for pre-trained LLM). For the pre-training LLM, we use the dataset collected in ~\citep{Padalkar2023OpenXR}, which contains 90000$\sim$ training scenes in the TAMP environment. The results in Table.~\ref{tab:RT1-TAMP} are tested when trained on $1\%$ of the dataset, corresponding to 300 examples for the two planning tasks. This setting is aligned with it in ~\cite{Driess2023PaLMEAE}. Most input representations exhibit comparable performance for 3-5 objects in the scene, equivalent to the quantity in the training set. ~\cite{Driess2023PaLMEAE} discussed the advantages of pre-training ViT with fully mixed data (text + visual), which will bring significant improvement. EMAC+ follows this pretraining approach. We found that although there is not much difference between EMAC+ and PaLM-E in visual QA, EMAC+ far outperforms PaLM-E in planning. This can be ascribed to EMAC+'s capacity to synchronize textual and visual states. In other words, both have employed pixel and textual observations during pretraining. During the training phase, EMAC+ can get deeper insights into environment dynamics by observing the execution of actions by the VLM agent and the LLM agent. This information is irreplaceable for action planning.

\begin{table}[tbh]
    \centering
    \footnotesize
    \vspace{-1em}
    \resizebox{.8\linewidth}{5.8em}{
    \begin{tabular}{lcccccccc}
    \toprule
    \hline \multirow{2}{*}{Agent} & \multirow{2}{*}{Object centric} & \multirow{2}{*}{LLM pre-train} & \multicolumn{4}{c}{Embodied VQA} & \multicolumn{2}{l}{Planning} \\\cline{4-7} \cline{8-9}
     & & & $q_1$ & $q_2$ & $q_3$ & $q_4$ & $p_1$ & $p_2$ \\
    \midrule SayCan & n/a  & \checkmark & - & - & - & - & 38.7 & 33.3 \\
    PaLI  & n/a &\checkmark & - & 0.0 & 0.0 & - & - & - \\
    InstructBLIP & n/a & \checkmark & 47.4& 46.2& 41.4& 40.5& 48.4 & 40.1\\
    PaLM-E (no VQA) & \checkmark & \checkmark & - & - & - & - & 71.9 & 75.1 \\
    PaLM-E (OSRT) & \checkmark & \checkmark & \textbf{99.7} & 98.2 & \textbf{100.0} & \textbf{93.7} & 82.5 & 76.2 \\
    PaLM-E (State) & \checkmark & \ding{55} & 99.4 & 89.8 & 90.3 & 88.3 & 45.0 & 46.1\\
    \midrule
    EMAC+ (Ours)& \checkmark & \checkmark & 98.2 & 98.4& \textbf{100.0}& 91.6 & \textbf{90.5} & \textbf{88.4}\\
    EMAC+ (abla)& \checkmark & \ding{55} & 92.2 & 90.4& 88.3 &  82.4 & 74.6 & 72.2\\
    EMAC+ (frozen LLM) & \checkmark &\checkmark & 88.4 & 82.9 & 82.5 & 79.4 & 81.6 & 75.3\\
    \hline
    \bottomrule
\end{tabular}}
    \vspace{-1em}
    \caption{\footnotesize\textbf{Results on RT-1 TAMP environment.} where data from TAMP constitutes only $1\%$ (i.e., 320 samples for $p_1, p_2$ each) of total training data size. PaLM-E (OSRT) despites using no large-scale data, provides the most effective input encodings for learning. PaLM-E (states) provides ground-truth object-centric information during training. The LLM agent in all PaLM-E baselines is frozen (following the settings in ~\cite{Driess2023PaLMEAE}).}
    \label{tab:RT1-TAMP}
    \vspace{-1em}
\end{table}

\begin{wrapfigure}[10]{r}{0.55\textwidth}
    \centering
    \footnotesize
    \vspace{-1.3em}
    \resizebox{.55\textwidth}{4.2em}{
    \begin{tabular}{lcccccc}
    \toprule
    \hline
    \multirow{2}{4em}{Agent} &  \multirow{2}{4em}{LLM+ViT Pretrain} & \multirow{2}{4em}{Task Finetune} & \multirow{2}{4em}{Frozen LLM} & \multicolumn{3}{c}{\textbf{Success Rate}} \\
    \cline{5-7} &&& & Task 1 & Task 2 & Task 3 \\
    \midrule 
    PaLM-E-12B   & \checkmark &\ding{55} &\checkmark & 20.0 & 2.5 & 11.3 \\
    PaLM-E-84B & \checkmark &\ding{55} &\ding{55} & \textbf{70.0} & 30.3 & 54.5 \\
    SayCan & n/a & \checkmark & \checkmark & 0.0 & - & - \\
    PaLI & \checkmark & \checkmark & \ding{55} & 0.0 & - & -\\
    \midrule
    EMAC+ (ours) &\checkmark &\checkmark &\ding{55} & 60.0 & \textbf{42.5} & \textbf{30.2} \\
    EMAC+ (abla-1) &\checkmark &\ding{55} &\checkmark & 20.0 & 10.2 & 12.4\\
    EMAC+ (abla-2) &\checkmark &\ding{55} & \ding{55} & 30.0 & 14.8 & 22.6 \\
    \hline
    \bottomrule
    \end{tabular}}
    \vspace{-1em}
    \caption{\footnotesize\textbf{Results on RT-1 planning tasks in the simulation.} EMAC+ (ours) denotes the results of our model, and EMAC+ (alba-1, abla-2) denotes the ablation studies.}
    \label{tab:RT1-planning}
\end{wrapfigure}

\textbf{RT-1 Planning tasks.} Figure.~\ref{tab:RT1-planning} reports the performance on long-horizon tasks from the Language-table environment~\citep{Lynch2022InteractiveLT}. Refer to Table.~\ref{tab:task_planning} for the task prompts in each planning task. The results are evaluated under a few-shot regime (10 demos per task) to assess the agent's generalization capability in out-of-distribution (OOD) tasks. Our findings indicate that EMAC+, possessing merely 7B parameters (Vicuna-7B), significantly surpasses PaLM-12B. The collaborative structure of EMAC+ significantly enhances the generalization capability of the LLM agent for unseen tasks, whereas SyaCan and PaLI are ineffective and incapable of addressing any test tasks.

\subsection{Ablation Studies}

\begin{figure}
    \centering
    \includegraphics[width=.9\linewidth]{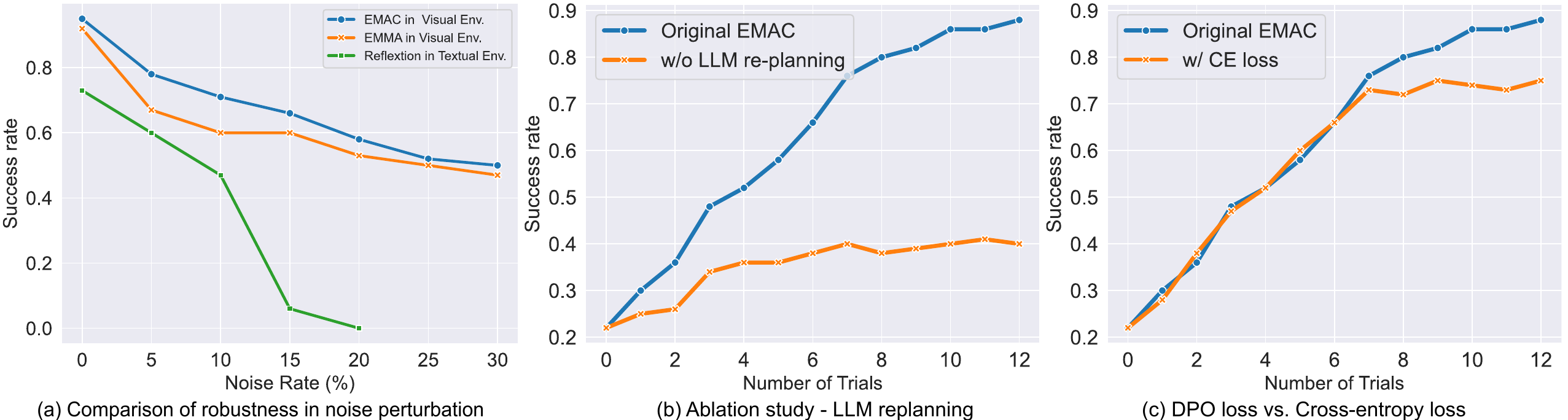}
    \vspace{-.5em}
    \caption{\footnotesize\textbf{Ablation Studies in ALFWorld.} Left: ``Comparison of robustness in noise perturbation''. To compare the robustness of EMAC+ and SOTA LLM agent (Reflexion), VLM agent (EMMA), we crop a random portion of the pixel observation with a specific noise rate. For LLM agent that can only interact with Textual world, we randomly replace some tokens in the textual observation with arbitrary ones. Mid: ``w/o LLM re-planning'' denotes the ablation study that removes the LLM finetune step by Eq.~\ref{equ:finetune_loss}. Right: ``w/ CE Loss'' denotes replacing DPO loss (Eq.~\ref{equ:loss}) with a token-level cross-entropy loss.}
    \label{fig:enter-label}
    \vspace{-1.5em}
\end{figure}

\textbf{LLM replanning can improve EMAC+'s performance over time.} In Figure.~\ref{fig:enter-label} (Mid), We restrict the LLM's capacity to modify its planned actions based on execution feedback, emulating a static planner similar to previous research (e.g., SayCan). The VLM performs as expected; however, the LLM fails to revise its strategy following failures. The results indicate that the static LLM redundantly executes impractical actions (e.g., attempting to grasp obscured objects) and fails to adapt to environment dynamics (e.g., item displacement caused by slippage). LLM replanning facilitates adaptive refinement through the internalization of environment dynamics. For instance, after unsuccessfully attempting to open a resistant drawer, our comprehensive model learns to suggest ``Apply upward force while pulling'' in future attempts. The ablated LLM, devoid of this input, continues to execute the first ``Pull the drawer'' action, resulting in repeated failures. This illustrates that closed-loop replanning is crucial for addressing open world uncertainty.

\begin{wrapfigure}[12]{r}{0.55\textwidth}
    \centering
    \footnotesize
    \vspace{-1.3em}
    \resizebox{.55\textwidth}{4.6em}{
    \begin{tabular}{lccccc}
    \toprule
    \hline
    \multirow{2}{4em}{Agent} &  \multirow{2}{4em}{LLM+ViT Pretrain} & \multirow{2}{4em}{Task Finetune} & \multirow{2}{4em}{Frozen LLM} & \multirow{2}{4em}{Failure Detection} & \multirow{2}{4em}{Affordance} \\\\
    \midrule 
    PaLM-E (single)   & \checkmark & \ding{55} & \checkmark & 0.91 & 0.78  \\
    PaLM-E (mixture) & \checkmark &\ding{55} &\ding{55} & 0.77 & 0.91  \\
    PaLM-E (mixture) & \checkmark &\ding{55} &\checkmark & 0.91 & 0.87  \\
    CLIP-FT-hindsight & \ding{55} & \checkmark & \checkmark & 0.89 & -  \\
    PaLI & \checkmark & \checkmark & \ding{55} & 0.73 & 0.62 \\
    \midrule
    EMAC+ (ours) &\checkmark &\checkmark &\ding{55} & 0.88 & 0.94 \\
    EMAC+ (abla-1) &\checkmark &\ding{55} &\checkmark & 0.70 & 0.64 \\
    EMAC+ (abla-2) &\checkmark &\ding{55} & \ding{55} & 0.72  & 0.59\\
    \hline
    \bottomrule
    \end{tabular}}
    \vspace{-1em}
    \caption{\footnotesize\textbf{Results on RT-1 Mobile Manipulation tasks in the simulation.} EMAC+ (ours) denotes the results of our model, and EMAC+ (abla-1, abla-2) denotes the ablation studies.}
    \label{tab:RT1-manipulation}
\end{wrapfigure}

\textbf{DPO loss vs. Cross-entropy loss.} In EMAC+, the collaborative LLM and VLM scheme is achieved through the DPO loss across the different action sequences. To evaluate the effectiveness of DPO loss, we conduct an ablation study to replace it with cross-entropy loss and report the results in Figure.~\ref{fig:enter-label}(Right). We observe that CE loss can deliver a comparable learning velocity to DPO loss during training, however fails to attain equivalent performance. CE Loss considers action tokens as autonomous predictions, optimizing the probability of each distinct token. This disregards time relationships among actions. Conversely, DPO loss enhances the complete action sequence as a preference hierarchy, because pathways that more effectively fulfill the LLM’s plan receive elevated rewards. This delineates causal links among steps.

\section{Conclusion and limitations}
In this work, we presented EMAC+, a novel embodied multimodal agent designed to collaboratively integrate large language models (LLMs) and vision-language models (VLMs). EMAC+ successfully addresses critical limitations of current static multimodal approaches by enabling a dynamic, bidirectional feedback loop between textual planning and visual control. Through rigorous evaluations on ALFWorld and RT-1 benchmarks, EMAC+ demonstrates superior performance in task accuracy, robustness, and learning efficiency. Our detailed ablation studies and practical considerations further underscore EMAC+'s advantages and applicability. Overall, EMAC+ paves a promising pathway towards more adaptive and intelligent embodied agents.

Although EMAC+ performs well across two benchmarks, our evaluation does not include real-world robotics tasks due to the limits of idle equipment. On the other hand, while EMAC+ significantly improves generalization and adaptability, challenges remain in scenarios involving highly dynamic and unpredictable physical interactions, highlighting opportunities for future research to enhance the model's resilience and generalizability.


\subsubsection*{Ethics Statement}
This work investigates embodied multimodal agents in simulated environments (ALFWorld and RT-1) and does not involve human subjects, personal data, or real-world robotic deployments. Therefore, there are no direct ethical risks in the experiments presented. We acknowledge, however, that embodied AI systems deployed in real-world settings could introduce safety risks, particularly if planning or perception errors occur. To mitigate such risks, we emphasize that EMAC+ is a research prototype evaluated in controlled simulations. We encourage future work to incorporate safety constraints, human oversight, and rigorous evaluation before applying such systems in physical environments.

\bibliography{ICLR_2026_submission/iclr2026_conference}
\bibliographystyle{iclr2026_conference}


\newpage
\appendix

\subsubsection*{Clarification for the use of LLM}
This work employs large language models (LLMs) strictly as a technical component of our proposed framework. Specifically, an LLM expert is integrated into EMAC+ to generate high-level task plans and refine them through feedback from a vision-language model (VLM). The LLM is not used for dataset generation, data annotation, or to assist in writing this paper. All training and evaluation rely on publicly available environments (ALFWorld and RT-1), and the role of the LLM is confined to agent planning within these benchmarks.
\section{Experiments Details for EMAC+} \label{sec:prompts}

\subsection{Evaluation metrics for manipulation tasks (Table.~\ref{tab:RT1-manipulation}).}

In RT-1 manipulation tasks, we mainly follow the setup in ~\cite{Ahn2022DoAI}, where the robot needs to plan a sequence of navigation and manipulation actions based on the task instruction and pixel observations. We have two test metrics to evaluate the performance, i.e., affordance prediction and failure detection. Affordance prediction is designed to investigate EMAC+'s performance on predicting whether a skill of the low-level policy can be executed in the environment. More specifically, given an \texttt{<image>} and a question ``\texttt{Does the control policy can do <skill> here?}''. Failure detection is commonly used for robotic closed-loop planning to detect failures. For example, given an \texttt{<image>}, the model needs to understand ``\texttt{Is <skill> successfully executed by the robot?}''. This is an important attribute of the multi-modal agent. It is theoretically essential for the agent to comprehend the reasons for the failure of the current task. It provides guarantee for the LLM's ability to modify plans dynamically.

\begin{wrapfigure}[13]{r}{.55\textwidth}
    \centering
    \renewcommand{\arraystretch}{1.2}
    \begin{tabular}{l}
    \toprule
    \hline
    \pbox{\linewidth}{\textbf{Task 1:} Q1: There is a block that is closest to \{i.e., top \\ right corner\}. Push that block to the other block of \\the same color.} \\
    \midrule
    \pbox{\linewidth}{\textbf{Task 2:} Q2: How to sort the blocks by colors into \\corners?}\\
    \midrule
    \pbox{\linewidth}{\textbf{Task 3:} Q3: How to push all the blocks that are on \\ the \{left/right\} side together, without bringing over \\ any of the blocks that are on the \{right/left\} side?}\\
    \hline
    \bottomrule
    \end{tabular}
    \caption{\footnotesize\textbf{Task prompts for planning tasks. (Figure.~\ref{tab:RT1-planning})}}
    \label{tab:task_planning}
\end{wrapfigure}
\subsection{Translated Textual state and LLM prompts in ALFWorld}
We supply all LLM prompts for the training process (Alg.~\ref{alg:design}) of EMAC+. We utilize the prompting strategy established by ReAct~\citep{Yao2022ReActSR} while disregarding the reasoning traces, specifically the "think" stages, throughout the imitation learning process between EMAC+ and the LLM actor. Following each trial $i$, the retrospective feedback $\mathcal P_i$ produced by the LLM critic will be incorporated into the long-term memory $\mathcal P$. In practice, we limit $\mathcal P$ to a maximum of 1-3 stored feedback instances to comply with the maximum context duration of the LLM. The textual state and task instruction is shown in Table.~\ref{tab:pro_1}. The LLM plans the action sequence to complete the task as shown in Table.~\ref{tab:pro_2}. The agent executes the planned actions in the environment, then the LLM generates the retrospective feedback based on the environment feedback (Table.~\ref{tab:pro_3}). The LLM expert will re-plann actions if the agent cannot successfully execute the previous actions (Table.~\ref{tab:pro_4}). 

\begin{table}[tbh]
    \centering
    \begin{tabular}{l}
    \toprule
    \pbox{0.9\linewidth}{\textbf{\texttt{Environment:}} \texttt{You are in the middle of a room. Looking quickly around you, you see a cabinet 4, a cabinet 3, a cabinet 2, a cabinet 1, a countertop 1, a garbagecan 1, a handtowelholder 2, a handtowelholder 1, a sinkbasin 2, a sinkbasin 1, a toilet 1, a toiletpaperhanger 1, and a towelholder 1.}} \\
    \textbf{\texttt{task instruction:}} \texttt{Your task is to: put some spraybottle on toilet.} \\
    \bottomrule
    \end{tabular}
    \caption{An example of the prompt for MAC+ in ALFWorld}
    \label{tab:pro_1}
\end{table}

\begin{table}[tbh]
    \centering
    \begin{tabular}{l}
    \toprule
    \pbox{0.9\linewidth}{\texttt{> think: To solve the task, I need to find and take a sparybottle, then put it on toilet.}} \\
    \pbox{0.9\linewidth}{\texttt{> think: First I need to find a spraybottle. A spraybottle is more likely to appear in cabinet (1-4), countertop (1), toilet (1), sinkbasin (1-2), garbagecan (1). I can check one by one, starting withcabinet 1.} }\\
    \textbf{\texttt{Action Sequence:}}\\
    \texttt{> step 1: go to cabinet 1} \\
    \texttt{> step 2: go to cabinet 2} \\
    \texttt{> step 3: take spraybottle 2 from cabinet 2} \\
    \texttt{> step 4: go to toilet 1} \\
    \texttt{> step 5: put spraybottle 2 in/on toilet 1}\\
    \bottomrule
    \end{tabular}
    \caption{Action sequence generated by the LLM}
    \label{tab:pro_2}
\end{table}

\begin{table}[tbh]
    \centering
    \begin{tabular}{l}
    \toprule
    \textbf{\texttt{Action Sequence:}}\\
    \texttt{> step 1: go to cabinet 1} \\
    \pbox{0.9\linewidth}{\texttt{\textbf{Env. feedback:} On the cabinet 1, there is cloth 1, a soapbar 1, a soapbottle 1}}\\
    \pbox{0.9\linewidth}{\texttt{> think: On the cabinet 1, I can see cloth 1, a soapbar 1, a soapbottle 1. I can continue to execute step 2.}}\\
    \texttt{> step 2: go to cabinet 2} \\
    \pbox{0.9\linewidth}{\textbf{\texttt{Env. feedback:}} \texttt{cabinet 2}}\\
    \pbox{0.9\linewidth}{\texttt{> think: cabinet 2 is closed. I can continue to execte step 3.}}\\
    \texttt{> step 3: take spraybottle 2 from cabinet 2} \\
    \pbox{0.9\linewidth}{\textbf{\texttt{Env. feedback:}} \texttt{[Action failed] cabinet 2.}}\\
    \\ 
    $\cdots$
    \\
    \texttt{Environment Return. Failed.} \\
    $\cdots$
    \\
    \textbf{\texttt{long-term memory}} $\mathcal P:$ \texttt{[$\cdots$]} \\
    \texttt{\textbf{Trajectory:} [target environment, task instruction, $\cdots$]}\\
    \bottomrule
    \end{tabular}
    \caption{Environment feedback}
    \label{tab:pro_3}
\end{table}

\begin{table}[tbh]
    \centering
    \footnotesize
    \begin{tabular}{l}
    \toprule
    \texttt{> Q: Was the planned action executed successfully?}\\
    \texttt{> No.}\\
    \pbox{0.9\linewidth}{\texttt{> think: step 3 is failed to execute. cabinet 2 is closed.}} \\
    \texttt{> Q: Does the current plan can solve the task?}\\
    \texttt{> No.}\\
    \texttt{> think: I need to change the action sequence. I need to open cabinet 2.}\\
    \textbf{\texttt{Replanned Action Sequence:}}\\
    \texttt{> step 1: go to cabinet 1} \\
    \texttt{> step 2: go to cabinet 2} \\
    \textbf{\texttt{> step 3: open cabinet 2}} \\
    \texttt{> step 4: take spraybottle 2 from cabinet 2} \\
    \texttt{> step 5: go to toilet 1} \\
    \texttt{> step 6: put spraybottle 2 in/on toilet 1}\\
    \\
    \textbf{\texttt{long-term memory}} $\mathcal P:$ \texttt{[$\cdots$]} \\
    \texttt{\textbf{Trajectory:} [target environment, task instruction, $\cdots$]}\\
    \bottomrule
    \end{tabular}
    \caption{Retrospecitve feedback from LLM for re-planning}
    \label{tab:pro_4}
\end{table}

\subsection{Prompts in RT-1}
In Table.~\ref{tab:rt1_task_list}, We list the training tasks for EMAC+ in RT-1. At the onset of the episode, each robot autonomously approaches its designated station and conveys to the operator the instruction to be demonstrated to the robot. To guarantee a balanced dataset and randomization of the scene, we developed a software module tasked with sampling the instructions for demonstration and randomizing the background setting. Each robot instructs the demonstration on how to randomize the scene and which instruction to exhibit. Demonstrations are conducted with direct line-of-sight between the operator and the robot utilizing two virtual reality remotes. We align remote controls with our policy action space to maintain the consistency of transition dynamics. The three-dimensional location and rotational displacements of the remote are correlated to the six-dimensional displacements of the robotic tool. The joystick's x, y coordinates correspond to a rotational angle and travel distance of the mobile base. We calculate and monitor trajectories towards the target poses derived from the joystick commands.

\begin{table}[tbh]
    \centering
    \resizebox{\linewidth}{4em}{
    \begin{tabular}{lcll}
    \toprule
    \hline 
    Skill  &  Count &  Description &  Example Instruction  \\
    \hline 
    Pick \texttt{Object} & 130 & Lift the object off the surface  &  \texttt{pick iced tea can}  \\
    Move \texttt{Object} Near \texttt{Object} & 337 & Move the first object near the second & \texttt{move pepsi can near rxbar blueberry}  \\
    Place \texttt{Object} Upright  & 8 & Place an elongated object upright  & \texttt{place water bottle upright } \\
    Knock \texttt{Object}Over  & 8 & Knock an elongated object over  & \texttt{knock redbull can over } \\
    Open / Close \texttt{Object} & 6 & Open or close any of the cabinet drawers  & \texttt{open the top drawer } \\
    Place \texttt{Object} into \texttt{Receptacle} & 84 & Place an object into a \texttt{receptacle} & \texttt{place brown chip bag into white bowl} \\
    \multirow{2}{12em}{Pick \texttt{Object} from \texttt{Receptacle}} & \multirow{2}{1em}{162} & \multirow{2}{16em}{Pick an object up from a location and then place it on the counter} & \multirow{2}{24em}{\texttt{pick green jalapeno chip bag from paper bowl and place on counter}}\\\\
    \midrule
    Total & 735 & \\
    \hline
    \bottomrule
    \end{tabular}}
    \caption{Training tasks for EMAC+ in RT-1}
    \label{tab:rt1_task_list}
\end{table}

\section{Additional Ablation Studies}
\begin{figure}[tbh]
    \centering
    \includegraphics[width=\linewidth]{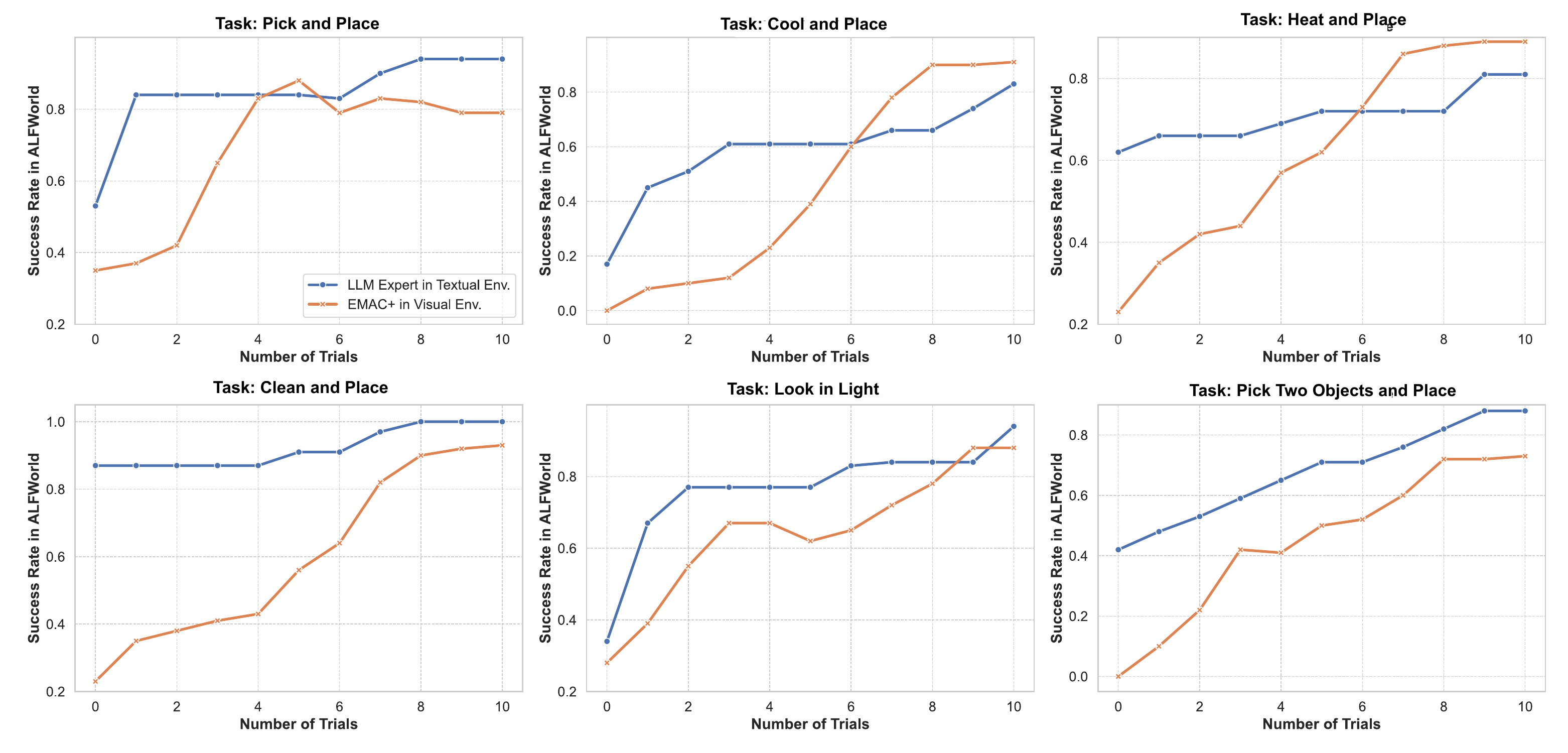}
    \caption{\textbf{Comparison of success rate between EMAC+ and the LLM expert.}}
    \label{fig:abla_llm_emac}
\end{figure}

\paragraph{EMAC+ learns knowledge from LLM expert.} In the early training stage of EMAC+, EMAC+ aligns its knowledge with an LLM expert by imitation learning. The LLM expert interacts with the textual environment and collects experience. As the number of trials increases, the LLM expert is able to improve its performance with long-term memory (retrospective feedback). We report the comparison of EMAC+ in the visual world and the LLM expert in the textual world in Fig.~\ref{fig:abla_llm_emac}. The results demonstrate that, under the guidance of the LLM expert, EMAC+ is able to effectively learn and execute tasks in visually grounded environments. Notably, EMAC+ eventually surpasses the performance of the LLM expert on several tasks. This performance gain highlights the advantage of grounding learning in the visual modality, which provides richer contextual and physical cues than text alone. These findings underscore the necessity of deploying multi-modal models in embodied environments, rather than relying solely on language-based agents, to fully exploit the informational richness of real-world sensory inputs.

\paragraph{LLM agent planning in the textual world.} In this section, we investigate how LLMs can deliver accurate planning in the textual world before getting corrective feedback from the domain expert. Table~\ref{tab:llm_trial_agent} presents the number of errors in the generated domain models. To help the readers understand the complexities of the action models, additionally report the total number of parameters and literals in the final corrected domain models produced by GPT-4v and Qwen-2.5-7B. We also report the performance of the LLM agents after receiving feedback from the textual engine. Despite their advanced language understanding capabilities, these models struggle to generate accurate and executable action plans without prior exposure, indicating a fundamental limitation in relying solely on language-based reasoning for complex embodied tasks. This observation underscores the critical role of EMAC+'s retrospective planning structure, which leverages environment feedback and multi-modal grounding to refine decision-making iteratively. The integration of such retrospective mechanisms is essential for enhancing robustness and generalization in multi-modal embodied agents.

\begin{table}[tbh]
    \centering
    \footnotesize
    \begin{tabular}{lcccc}
    \toprule
    \hline
    \multirow{2}{6em}{LLM agent} & \multirow{2}{8em}{Number of Errors (unseen)} & \multirow{2}{6em}{Number of Trials (unseen)} & \multirow{2}{6em}{Number of Errors (seen)} & \multirow{2}{6em}{Number of Trails (seen)}\\\\
    \midrule
    GPT-4v & 94 & 134 & 34 & 140\\
    GPT-3.5-Turbo & 106 & 134 & 46 & 140\\
    Qwen-2.5 & 120 & 134 & 52 & 140\\
    \hline
    \bottomrule
    \end{tabular}
    \caption{The number of errors in the action planning produced by the LLMs.}
    \label{tab:llm_trial_agent}
\end{table}

\section{The structure of EMAC+}
We list the structure of EMAC+ and hyperparameters in Table.\ref{tab:hyper_alf} and Table.~\ref{tab:hyper_rt1}. These hyperparameters are primarily based on those suggested for fine-tuning the InstructBLIP model~\citep{Dai2023InstructBLIPTG}. During training, we exclusively change the parameters of the linear projection layer while maintaining the remaining components in a frozen state, consistent with the approach utilized in instruction tuning for numerous existing studies~\citep{Gao2023LLaMAAdapterVP, Zhu2023MiniGPT4EV}. We employ the AdamW optimizer, initiating with a linear warmup of the learning rate, succeeded by a linear decay, culminating in a minimal learning rate of 0. Furthermore, we eliminate the instruction input of Q-Former, utilized in InstructBLIP, and observe an enhancement in performance across all studies.

\begin{table}[!hp]
    \centering
    \begin{tabular}{lc}
    \toprule
    \hline
    Hyperparameter & Value \\
    \hline \multicolumn{2}{c}{ EMAC's Architecture } \\
    LLM decoder & Vicuna-7b-v1.1~\cite{Zheng2023JudgingLW} \\
    Image encoder & ViT-L ~\cite{Radford2021LearningTV} \\
    Q-Former & BERT $_{\text {base }}$~\cite{Devlin2019BERTPO} \\
    Pretrained weights & InstructBLIP ~\cite{Dai2023InstructBLIPTG}\\
    Number of query tokens & 32 \\
    Q-Former text input & False \\
    Max text length & 1024 \\
    Image resolution & 224 \\
    \hline \multicolumn{2}{c}{Initialize $\pi_\theta$ to $\pi_{ref}$} \\
    Finetuning epochs & 6\\
    Warmup steps & 1000 \\
    Learning rate & $10^{-5}$ \\
    Batch size & 128 \\
    AdamW $\beta$ & $(0.9,0.999)$ \\
    Weight decay &  0.05 \\
    Drop path & 0\\
    Inference beam size & 5 \\
    \hline \multicolumn{2}{c}{Imitation Learning} \\
    Base model for LLM expert & LLaMa3 \\
    Prompts for LLM expert & refer to Sec.~\ref{sec:prompts} \\
    Number of trials & 12 \\
    Episode length & 30 \\
    Size of long-term memory & 3 \\
    Learning rate & $5 \times 10^{-6}$ \\
    Warmup steps & 300\\
    Batch size & 16 \\
    Training epochs per trial & 5 \\
    DPO $\beta$ & 0.1 \\
    \hline
    \multicolumn{2}{c}{LoRA Finetune}\\
    Optimizer & AdamW\\
    Warmup steps & 12 \\
    LR Schedule & Linear \\
    Batch Size & 20 \\
    Epoch & 40 \\
    Learning rate & $2*10^{-4}$\\
    Adaption specific & $r_q = r_v =8$\\
    \hline
    \bottomrule
\end{tabular}
    \caption{Hyperparameters of EMAC+ for ALFWorld experiments}
    \label{tab:hyper_alf}
\end{table}

\begin{table}[!hp]
    \centering
    \begin{tabular}{lc}
    \toprule
    \hline
    Hyperparameter & Value \\
    \hline \multicolumn{2}{c}{ EMAC's Architecture } \\
    LLM decoder & Vicuna-7b-v1.1~\cite{Zheng2023JudgingLW} \\
    Image encoder & ViT-L ~\cite{Radford2021LearningTV} \\
    Q-Former & BERT $_{\text {base }}$~\cite{Devlin2019BERTPO} \\
    Pretrained weights & InstructBLIP ~\cite{Dai2023InstructBLIPTG}\\
    Number of query tokens & 32 \\
    Q-Former text input & False \\
    Max text length & 1024 \\
    Image resolution & 224 \\
    \hline \multicolumn{2}{c}{Initialize $\pi_\theta$ to $\pi_{ref}$} \\
    Finetuning epochs & 6\\
    Warmup steps & 2500 \\
    Learning rate & $10^{-5}$ \\
    Batch size & 128 \\
    AdamW $\beta$ & $(0.09,0.999)$ \\
    Weight decay &  0.05 \\
    Drop path & 0\\
    Inference beam size & 5 \\
    \hline \multicolumn{2}{c}{Imitation Learning} \\
    Base model for LLM expert & LLaMa3 \\
    Prompts for LLM expert & refer to Sec.~\ref{sec:prompts} \\
    Number of trials & 8 \\
    Episode length & 12 \\
    Size of long-term memory & 3 \\
    Learning rate & $4 \times 10^{-6}$ \\
    Warmup steps & 120\\
    Batch size & 24 \\
    Training epochs per trial & 3 \\
    DPO $\beta$ & 0.1 \\
    \hline
    \multicolumn{2}{c}{LoRA Finetune}\\
    Optimizer & AdamW\\
    Warmup steps & 48 \\
    LR Schedule & Linear \\
    LR decay schedule &  linear decay to zero \\
    Batch Size & 256 \\
    Learning rate & $1*10^{-4}$\\
    Adaption specific & $r_q = r_v =8$\\
    \hline
    \bottomrule
\end{tabular}
    \caption{Hyperparameters of EMAC+ for RT-1 experiments}
    \label{tab:hyper_rt1}
\end{table}

\end{document}